\documentclass[10pt, a4paper]{article}

\usepackage{lrec-coling2024} % this is the new style

\usepackage{hyperref}
\usepackage{times}
\usepackage{latexsym}
\usepackage{soul}
\usepackage{microtype}

\usepackage{enumitem}

\usepackage{caption}
\usepackage{subcaption}
\usepackage{adjustbox}
\usepackage{multirow}

\title{Intent Detection and Entity Extraction from Biomedical Literature}

\name{Ankan Mullick\textsuperscript{1}*, Mukur Gupta\textsuperscript{2}*, \thanks{*Authors contributed equally}
Pawan Goyal\textsuperscript{1}} 

\address{\textsuperscript{1}Computer Science and Engineering Department, IIT Kharagpur, India \\\textsuperscript{2}Computer Science Department, Columbia University, USA\\
         ankanm@kgpian.iitkgp.ac.in, mukur.gupta@columbia.edu, pawang@cse.iitkgp.ac.in\\}

\abstract{
Biomedical queries have become increasingly prevalent in web searches, reflecting the growing interest in accessing biomedical literature. Despite recent research on large-language models (LLMs) motivated by endeavors to attain generalized intelligence, their efficacy in replacing task and domain-specific natural language understanding approaches remains questionable. In this paper, we address this question by conducting a comprehensive empirical evaluation of intent detection and named entity recognition (NER) tasks from biomedical text. We show that Supervised Fine Tuned approaches are still relevant and more effective than general-purpose LLMs. Biomedical transformer models such as PubMedBERT can surpass ChatGPT on NER task with only 5 supervised examples. 
}

\begin{document}

\maketitleabstract

\section{Introduction}
Research on large-language models has sky-rocketed in the post-ChatGPT era. Researchers are now aiming for generalized intelligence by increasing model size \cite{brown2020gpt3,chowdhery2022palm,hoffmann2022_scaling_laws}, expanding \& rearranging pretraining data \cite{touvron2023llama, touvron2023llama2, sarkar2021tso} and incorporating human feedback \cite{ouyang2022_instructgpt, dubois2023alpacafarm}. It is shown that the adoption of GPT-4 \cite{openai2023gpt4} can potentially affect up to 80\% of the U.S. workforce \cite{eloundou2023gptsaregpts}. These generalization reasoning demonstrations raise an important question for the research community - does this mark an end to the task and domain-specific natural language understanding approaches? While some research places LLMs as ``General Purpose Technologies" \cite{eloundou2023gptsaregpts, zhang2023small} for solving a range of complicated tasks, we show that these models struggle to perform well on domain-specific complex tasks and specialized Supervised Fine-tuned (SFT) models are still needed to solve language understanding use-cases.

\begin{figure}[t]
     \centering
     \begin{subfigure}[b]{0.49\columnwidth}
         \centering
         \includegraphics[width=\columnwidth]{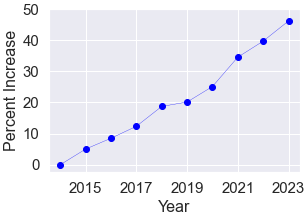}
         \caption{Google yearly Trend}
         \label{fig1-1}
     \end{subfigure}
     \hfill
     \begin{subfigure}[b]{0.49\columnwidth}
         \centering
         \includegraphics[width=\columnwidth]{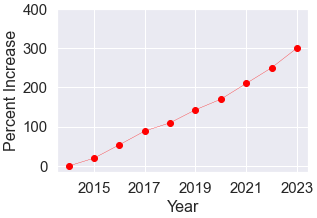}
         \caption{Bing Query yearly}
         \label{fig1-2}
     \end{subfigure}
     \hfill
     %\vspace{-6mm}
        \caption{Biomedical query search Statistics}
        %\vspace{-5mm}
        \label{fig-trends}
\end{figure}

Over the past two decades, web searches have evolved dramatically transitioning from generic interfaces to more intent-specific and entity-aware systems capable of immediately displaying diverse multi-modal responses. Particularly, biomedical inquiries, spanning topics such as medical treatment, medical diagnosis, disease, etc. have seen a surge in popularity across search engines. Fig. \ref{fig-trends} shows the increase in the percentage of Biomedical queries on Bing search %\footnote{We take the Bing query log for the 10$^{th}$ day of January of every year and filter this list using patterns for list-type queries. Then, we randomly select 200 queries out of this filtered set. We manually label the queries in this sample as biomedical queries or not. The graph shows the increase (in \%) in the fraction of queries meeting the filter criteria among all the queries in our random sample. It is significant for Bing.}
and Google trends\footnote{Google trends data of last 10 years on five topics (Health, Medical Treatment, Medical diagnosis, Disease, Pharmaceutical drug) was gathered from Google Trends (https://trends.google.com/trends/)}.
%\footnote{https://github.com/psgrghvuo/nlu-biomedical/tree/main} in the last decade.

As large volumes of biomedical data continue to be generated every second on various online platforms
%\footnote{\url{https://tinyurl.com/telegraph-news}, %https://www.ncbi.nlm.nih.gov/research/coronavirus/, 
%\url{https://tinyurl.com/bio-lit}}, 
the role of information retrieval systems in processing domain-specific texts becomes increasingly important. However, handling biomedical text data presents unique challenges, as the medical queries on search engines and online medical forums are often incomplete, do not follow a specific structure, and contain hard-to-interpret context-specific medical terminologies, as shown in Table \ref{cmid-id-examples}. While recent research is centered around the development of general-purpose LLMs, that are shown to exhibit exceptional Common Sense Reasoning capabilities \cite{touvron2023llama2}, we show that these models face challenges in transferring their performance to intricate biomedical domains. %In addressing this gap, we show that domain-specific Supervised Fine-Tuned (SFT) approaches outperform LLMs in two crucial natural language understanding tasks on biomedical domain.
To this end, we focus on two crucial natural language understanding tasks of intent detection and named entity recognition from biomedical text.

%Most of the research is centered around building generic domain-oriented intent detection and entity extraction models to process the domain area-specific texts better and provide better responses. 
%Two different types of intents (Drug and Disease category) along with corresponding entities (Drug name, Disease Name, or condition) are shown in Table \ref{cmid-id-examples} (colored).

%With the big volumes of data being generated every second on the internet, the role of advanced Information Retrieval (IR) systems becomes increasingly important. As the first stage of such IR systems, lies the Natural Language Understanding systems capable of identifying the user query intent and/or locating the key entities in the text. These tasks becomes very difficult in the case of biomedical text data. This is because the medical queries on the online medical forums are often incomplete ,do not follow a specific structure and contain hard to interpret medical terminologies. Intent detection examples in Table \ref{cmid-id-examples} shows unstructured medical queries. Some examples showing hard to understand medical terms in NER are shown in \ref{ncbi-ner-examples}.\\

\begin{table}[h]
\centering
%\vspace{-1.5mm}
\begin{adjustbox}{width=0.95\columnwidth}
\begin{tabular}{p{0.9\columnwidth}  p{0.1\columnwidth}}
%\begin{tabular}{lll}
\hline
\textbf{Biomedical Text} & \textbf{Intent} \\
\hline
Pharmacokinetic properties of \colorbox{lime}{abacavir} were not altered by the addition of either \colorbox{lime}{lamivudine} or \colorbox{lime}{zidovudine} or the combination of \colorbox{lime}{lamivudine} and \colorbox{lime}{zidovudine}. & \colorbox{lime}{Drug}\\
\\
\hl{Canavan disease}, or \hl{spongy degeneration of the brain}, is a severe \hl{leukodystrophy} caused by \hl{deficiency of aspartoacylase (ASPA)}. & \hl{Disease}\\
\hline
\end{tabular}
\end{adjustbox}
%\vspace{-2mm}
\caption{Intent \& corresponding Entity (highlighted) examples from DDI and NCBI Disease datasets.}
\label{cmid-id-examples}
%\vspace{-3.5mm}
\end{table}

For the past two decades, different directions of intent detection and corresponding entity extraction have been explored. \cite{sun2016online,wang2020active,mu2017streaming,mu2017classification} demonstrate intent detection in the form of out-of-domain data detection. Other research works explore methods like few shot \citep{xia2021incremental}, zero-shot \cite{xia2018zero}, and clustering frameworks \cite{mullick2022framework}. %\cite{yani2022better,sufi2021automated,zhao2021bert, xlmr-2022-naacl} 
\cite{yani2022better,zhao2021bert, xlmr-2022-naacl}  explore entity recognition task in various settings. 
%However, most of these approaches are very domain-specific and hence not very effective for the biomedical domain. 
In the medical domain, \citet{zhou2021natural} focuses on smart healthcare % including hospital management, personal care, public health, etc. %\citet{bao2020hhh} propose hierarchical Bi-LSTM attention layer based work on question-answering system to build a chat-bot framework using user intents. 
and \cite{increasing-biomed-Exploiting-and-assesing, increasing-biomed-towards-reliable, biobert} inspect transformer based models for biomedical literature.
\citet{mullick2023intent, mullick2022fine, mullick2023novel, mullickexploring} aims at intent detection and entity extraction and % in Indic languages and %\citet{chinaei2014dialogue} focus on healthcare dialogue intents from user data. 
 %\citet{razzaq2017intent,amato2017chatbots} develop an e-Health application using intent-context relationships.  %\citet{amato2017chatbots} explore intents to build eHealth applications. 
 \citet{zhang2017bringing} explore medical query intents by applying graph-based frameworks. 
 \cite{mullick2024matscire,guha2021matscie} work on domain specific entity and corresponding relation extraction. \cite{mullick2017generic, mullick2016graphical, mullick2018identifying, mullick2018harnessing, mullick2019d,mullick2017extracting} aim at opinion-fact entity extraction.
 %to understand semantic structure of user intents in a medical query. 
 %\cite{xlmr-2022-naacl} build XLM-R based approach %to jointly integrate both textual and gazetteer information dynamically in few shot setting 
 %for entity extraction in multi-lingual settings.
 
 %Most of the works are in Chinese language and there is no generic universal architecture for intent detection and entity extraction across various datasets in biomedical literature. 
 There is no unified and exhaustive comparison of existing approaches with the recent LLMs for intent detection and entity extraction tasks across various datasets in biomedical literature.
 Our work differs from the prior research in two ways: we present a thorough empirical evaluation of the intent detection on three datasets and corresponding named entity extraction (NER) approaches on 27 unique entities covered in 5 biomedical datasets spanning across domains like drugs, diseases, chemicals, genetics and, human anatomy. 
 We evaluate various supervised approaches (transformer-based, handcrafted features, etc.) and benchmark them against two widely used large language models in the biomedical domain.
 %We evaluate different approaches (transformer, large language model, etc.) and provide two frameworks for two tasks in the biomedical domain. 
 Our experiments reveal that the biomedical transformer-based PubMedBERT model outperforms few-shot prompted ChatGPT (Turbo 3.5) on 4 biomedical NER benchmarks with just 5 supervised examples.
 We make our code publicly available.\footnote{\href{https://github.com/bioNLU-coling2024/biomed-NER-intent_detection}{https://github.com/bioNLU-coling2024/biomed-NER-intent\_detection}}
%We show that large language models like ChatGPT (Turbo 3.5), and Llama-2 fail to achieve similar performances to other supervised finetuned models for two domain-specific tasks - intent detection and entity extraction from the biomedical domain.

\section{Datasets}

We show our comparative study on a variety of datasets, which are widely used as benchmarks in the biomedical domain. We use five different Named Entity Recognition datasets:
JNLPBA \cite{jnlpba-corpus}, 
DDI (combining DDI-Drugbank and DDI-Medline) \cite{drug}, 
BC5CDR \cite{bc2gm-corpus}, 
NCBI-Disease \cite{bc5cdr-corpus} and
AnatEM \cite{anem-corpus}. 
Dataset statistics including the entity types, count, and train-test splits are outlined in Table \ref{tab:entity-data}. We use the pre-defined train-test divisions from the respective manuscripts. 

Along with the two popular intent detection datasets - CMID \citep{cmid} and KUAKE-QIC (part of the CBLUE \citep{cblue} benchmark), we combine the three of the above five NER datasets (JNLPBA, DDI, and NCBI-Disease) with respective intent labels (DDI for drugs, NCBI-Disease for disease and JNLPBA for Genetics) for intent classification task - termed as ``Intent-Merged'' dataset. Dataset statistics are summarized in Table \ref{tab:intent-data}.

CMID and KUAKE-QIC datasets, which are originally in Chinese, are translated to English using Google Translation API. For translation validation, a random sample of 400 translated (to English) examples of each dataset are validated manually by two Chinese experts (ALA Language Center Company) with HSK Level-3 proficiency. The human-validation shows 91.75\% and 97.0\% translation accuracy for CMID and KUAKE-QIC respectively. Hence, we use the translated English data along with their pre-defined intent labels for our experiments. The inter-annotator agreement is 0.89.

\begin{table}[t]
\centering
\begin{adjustbox}{width=0.5\textwidth}
\begin{tabular}{ccccc}
\hline
\textbf{Dataset}     &  \textbf{Entity Type} & \textbf{\# Entities} &   \textbf{\#Train}   &   \textbf{\#Test}\\
\hline
JNLPBA  &  Gene \& Protein & 5  & 2000        &404 \\
DDI  &  Drug & 4   & 714       &112 \\
BC5CDR& Chem \& Diesease & 2   &1000      &500 \\
NCBI-Disease& Disease & 4  &693      &100\\
AnatEM& Anatomy & 12  &300      &200\\
\hline
\end{tabular}
\end{adjustbox}
%\vspace{-1.5mm}
\caption{Statistics of the NER datasets. We use the pre-defined train-test split as mentioned in the papers.}%\ankan{who did this division? predefined? why it is not uniform}}
%\vspace{-4mm}
\label{tab:entity-data}
\end{table}

\begin{table}[!h]
%\vspace{-2mm}
\centering
\begin{adjustbox}{width=0.8\columnwidth}
\begin{tabular}{cccc}
\hline
        \textbf{Dataset} & \textbf{\#Train}  & \textbf{\#Test Size} &\textbf{\#Intents}\\
        \hline
        CMID & 9558 & 2696 &4\\
        KUAKE-QIC & 6931 & 1955 &11\\
        Intent-Merged & 3905 & 909 &3\\
        \hline
\end{tabular}
\end{adjustbox}
%\vspace{-2.5mm}
\caption{Statistics of Intent Detection datasets.}
%\vspace{-6mm}
\label{tab:intent-data}
\end{table}

\section{Experimental Settings}

\subsection{Intent Detection}
Intent detection is a multi-class classification task where we evaluate the accuracy of instruction-tuned %LLaMa-2 (Llama-2-7b-chat) and 
ChatGPT (gpt-3.5-turbo-instruct) against various SFT models on three English datasets: CMID, KUAKE-QIC, and Intent-Merged.

\noindent \textbf{1. Large Language Models:} 
To ensure consistency with prior works, we employ a $k$-shot prompt design, wherein $k$ examples per class from the training set are used in the prompt. Given the larger text sizes of the Intent-Merged dataset and the limited context window of LLMs, we use $k=1$ for all datasets. We note no significant performance improvement with increasing $k$ for CMID and KUAKE-QIC datasets. Further details on the prompt template are included in the GitHub repository.

\noindent \textbf{2. Supervised Fined-Tuned Models:} For SFT, we finetune - BERT (bert-base-uncased) \cite{bert}, RoBERTa (roberta-base) \cite{liu2019roberta}, PubMedBERT \cite{pubmedbert}, FastText \cite{cmid} and TextCNN~\cite{kim2014convolutional}. \\The empirical evaluation is shown in Table \ref{experiments-intent}.

\begin{table}[!h]
%\vspace{-1mm}
\centering
\begin{adjustbox}{width=0.95\columnwidth}
\begin{tabular}{ccccc}
\hline
        \textbf{Model} & \textbf{CMID}  & \textbf{KUAKE-QIC} & \textbf{Intent-Merged} & \textbf{Mean}\\
        \hline
        BERT & 72.26 & 75.91 & 96.37 & 81.51\\
        RoBERTa & \textbf{72.88} & \textbf{78.16} & \textbf{99.11} & \textbf{83.38}\\
        PubMedBERT & 72.70 & 76.88 & 97.90 & 82.49\\
        Llama-2 & 51.11 & 42.50 & 39.54 & 44.38\\
        ChatGPT & 42.36 & 44.04 & 64.44 & 50.28\\
        Fasttext & 68.43 & 72.48 & 96.80 & 79.24\\
        TextCNN & 70.69 & 75.19 & 96.15 & 80.68\\
        \hline
\end{tabular}
\end{adjustbox}
%\vspace{-1mm}
\caption{Accuracy (in \%) of intent classification tasks on three datasets.}
\label{experiments-intent}
%\vspace{-3mm}
\end{table}

All the SFT approaches consistently outperform the instruction-tuned %vanilla LLaMa-2 and 
ChatGPT. The poor performance of LLMs on the Intent-Merged dataset, which is quite easy for all the SFT approaches, reflects their deficiency in domain-specific knowledge within their general-purpose pretraining datasets. This also shows that models like FastText can outperform ChatGPT, given domain-specific finetuning. We note that transformer architectures give better performance on the translated corpus compared to FastText and TextCNN, which are shown to work well on Chinese data~\cite{cmid}. RoBERTa gives the highest accuracy across overall mean and individual datasets.

\begin{table*}[htb]
%\vspace{-5mm}
\centering
\begin{adjustbox}{width=0.92\textwidth}
\begin{tabular}{l|ccccc|c}
            \hline

\textbf{Model/Dataset} & \textbf{DDI}  & \textbf{JNLPBA} & \textbf{BC5CDR} & \textbf{NCBI Disease} & \textbf{AnatEM} & \textbf{Mean}\\
\hline
BERT (A)    &$83.94\pm 0.17$       &$72.60\pm 0.13$      &$87.13\pm 0.24$     &$77.44\pm 0.75$      &$78.66\pm 0.35$  &79.95\\
RoBERTa (A)  &$87.13\pm 0.44$       &$74.91\pm 0.11$      &$89.50\pm 0.09$     &$81.67\pm 0.51$      &$81.90\pm 0.60$   &83.06\\
\hline
BioBERT (B) &$88.06 \pm 0.08$       &$74.02 \pm 0.40$      &$90.19\pm 0.13$     &$81.91\pm 0.80$     &$83.43\pm 0.36$ &83.52\\
PubMedBERT (B)  & $88.84\pm 0.12$       & $75.15\pm 0.06$      & $90.77\pm 0.08$     & $82.34\pm 0.16$    & $84.21\pm 0.21$   &84.26\\
BioMed RoBERTa (B)  &$88.76\pm 0.31$       &$75.14\pm 0.25$      &$90.24\pm 0.18$     &$82.13\pm 0.92$     &$82.70\pm 0.17$    &83.80\\
Clinical BERT (B) &$83.79\pm 0.22$       &$72.54 \pm 0.07$      &$87.90 \pm 0.17$     &$76.34  \pm 0.64$     &$73.47 \pm 0.53$   &78.80\\
\hline
LSTM (C) &$73.00\pm 0.01$       &$67.00\pm 0.01$      &$79.01\pm 0.01$     &$70.00\pm 0.01$      &$74.01\pm 0.01$    &72.60\\
LSTM + CRF (C) &$74.75\pm 0.04$       &$70.67 \pm 0.01$      &$80.47 \pm 0.02$     &$73.13 \pm 0.02$     &$77.39 \pm 0.05$    &75.23\\
CNN (C) &$73.07\pm 0.01$       &$68.04\pm 0.01$      &$80.00\pm 0.01$     &$67.09\pm 0.01$      &$72.08\pm 0.01$    &72.06\\
CNN + CRF (C) &$73.29\pm 0.08$       &$70.84\pm 0.11$      &$81.27\pm 0.14$     &$73.59\pm 0.09$      &$75.15\pm 0.13$    &74.83\\
\hline
Logistic Regression (D) &$78.63\pm 0.01$       &$57.03 \pm 0.01$      &$78.20 \pm 0.01$     &$56.36  \pm 0.01$     &$65.48 \pm 0.01$   &67.14\\
XGBoost (D) &$73.55\pm 0.01$       &$53.06 \pm 0.01$      &$67.86 \pm 0.01$     &$52.62  \pm 0.01$     &$59.91 \pm 0.01$   &61.40\\
\hline
BINDER-BioBERT (E) &$89.01 \pm 0.01$      &$76.63 \pm 0.19$      &$91.59 \pm 0.09$     &$\textbf{85.47} \pm 0.36$     &$86.71 \pm 0.25$   & 85.88\\
\textbf{BINDER-PubMedBERT} (E) &$\textbf{89.12} \pm 0.01$       &$77.01 \pm 0.01$      &$\textbf{91.88} \pm 0.01$     &$85.25 \pm 0.02$     & $\textbf{86.95} \pm 0.02$   &\textbf{86.04}\\
BINDER-RoBERTa (E) &$87.98 \pm 0.01$       &$\textbf{77.08} \pm 0.01$      &$90.48 \pm 0.03$     &$84.62 \pm 0.06$     &$83.91 \pm 0.05$   &84.81\\
\hline
%LLaMa-2 (E) &$8.12 \pm 0.0$       &$5.87 \pm 0.0$      &$16.57 \pm 0.0$     &$4.5 \pm 0.0$     &$0.9 \pm 0.0$   &7.19\\
ChatGPT (F) &$42.94 \pm 3.10$       &$24.5 \pm 1.89$      &$44.68 \pm 2.78$     &$19.65 \pm 1.21$     &$2.92 \pm 0.07$   &26.94\\
%\hline
%\cite{xlmr-2022-naacl} (F) &$84.90\pm 3.32$       &$74.91 \pm 0.04$      &$88.65 \pm 0.29$     &$80.72  \pm 0.27$     &$78.19 \pm 0.78$   &81.48\\
    \hline
    \end{tabular}
\end{adjustbox} 
%\vspace{-2mm}
\caption{Experiment results (Macro average F1-Scores and corresponding standard deviations) on different NER systems trained/finetuned and tested on 5 biomedical Datasets.} %Macro average F1-Scores and corresponding standard deviations from five experimental runs are reported.} %\textcolor{red}{correction-edit}}
%\vspace{-2mm}
\label{results-table}
\end{table*}

\iffalse
\begin{table}[!h]
\centering
\begin{adjustbox}{width=0.9\columnwidth}
\begin{tabular}{lllll}
\hline
        Task/ & \multicolumn{2}{c}{NER} & \multicolumn{2}{c}{Intent Detection}\\
        Model & WebMD  & 1mg & WebMD  & 1mg\\
        \hline
        RoBERTa & \textbf{77.35} & \textbf{90.57} & 79.59 & \textbf{84.83}\\
        PubMedBERT & 76.65 & 89.92 & 81.32 & 84.82 \\
        %Joint BERT & 71.42 & 88.00 & 77.17 & \textbf{85.71} \\
        Joint BioBERT & 76.17 & 88.57 & \textbf{82.15} & 84.82\\
        Joint PubMedBERT & 75.78 & 85.51 & \textbf{82.15} & 83.03\\
        \hline
\end{tabular}
\end{adjustbox}
\caption{F1-Score of Joint Intent and Entity Extraction for two IHQID datasets \cite{mullick2023intent}}
\label{experiments-joint-ner}
\end{table}
\fi

\subsection{Named Entity Recognition}

For NER, we apply a strict match between the predicted entity class and the entity word boundaries and report strict F1-score (as in CoNLL shared task~\citep{conll2003}). 
%\mukur{Most of earlier works \cite{binder, pubmedbert, biobert} in this domain report exact F1 instead of strict F1.}
%We create five different train-test split (80:20) for each five datasets, across all above models to perform the experiment five times and report the macro average F1-score along with the standard deviation (for each dataset). 
We run all models 5 times with different random initialization and report micro-average F1-score along with standard deviations. We also report the overall mean for each approach. For a fair comparison, a maximum sequence length of 512 tokens was used for all models, hence the texts larger the token length were further broken into multiple texts.

%\noindent \textbf{1. Supervised Fine-Tuned Models:} 
\subsubsection{Supervised Fine-Tuned Models} 
We thoroughly examine five different settings on the five biomedical datasets.

\noindent \textbf{Setting A:} Fine-tuned BERT and RoBERTa models are used (pre-trained on general English corpus) without domain pretraining.

\noindent \textbf{Setting B:} Transformer systems with continued pretraining on biomedical text. We fine-tune BioBERT, PubMedBERT, BioMed RoBERTa, and ClinicalBERT.

\noindent \textbf{Setting C:} LSTM and Convolutional Neural Networks (with/without CRF) are used to generate the word embeddings and softmax classifiers for tag prediction.
    %Models are trained with early stopping if the performance saturates, along with a learning rate of 0.015 on SGD optimizer. A dropout layer with dropout of 0.5 is used on top LSTM layer.
    %Neural Word Embedding with conditional random field: Similar embeddings from the above category are used, but instead of a softmax classifier, a Conditional Random Field layer is used for sequence labeling. 
    
\noindent \textbf{Setting D:} Hand-crafted word level features with ML classifier: (i) POS tag (ii) shallow parsing features like chunk tag (iii) orthographic boolean features like all capital, is alphanumeric, etc. (iv) n-gram features, etc. %(v) morphological features like prefix and suffix %(vi) contextual features like the feature set of the previous/next word. 
 We use the GENIA tagger\footnote{http://www.nactem.ac.uk/GENIA/tagger/} for POS and Chunk tag extraction. We apply XGBoost and a multi-label logistic regression model for NER tag prediction. 
 %\mukur{\noindent \textbf{E) Setting 5:} BINDER \cite{binder} along with domain specific (PubMedBERT and BioBERT) and non-domain specific (RoBERTa) encoders.}

\noindent\textbf{Setting E:} We use state-of-the-art NER model BINDER \cite{binder} along with domain-specific (PubMedBERT and BioBERT) and RoBERTa encoders.

%\noindent \textbf{2. Large Language Models:} 
\subsubsection{Large Language Models} 
We use instruction-tuned ChatGPT (gpt-3.5-turbo-instruct). 

\noindent \textbf{Setting F:} We modify \cite{wang2023gptner} for reconditioning NER as a Tag generation problem. In addition to the prompt design proposed by \cite{wang2023gptner}, following \cite{binder} we also add a short description for the entity. Each prompt infers only a single entity tag. Hence, each text instance is passed multiple times for tagging all the entity types. We provide two examples from the train set in each prompt. To motivate both high recall and prevent hallucination in Entity identification, we specifically pick examples with the median number of entity tags in the training dataset.

\iffalse{}
\section{Results}

\subsubsection{Intent Detection}
All the SFT approaches outperform the instruction-tuned LLaMa-2 and ChatGPT. Even the poor performance on the Intent-Merged dataset, which is quite easy for all the SFT approaches, reflects that LLMs don't have enough domain-specific knowledge in their general-purpose pretraining datasets. This also shows that very cheaper models (like FastText) can easily outperform massive 7 billion parametric LLaMa-2, given domain-specific finetuning. We note that transformer architectures give better performance on the translated corpus than FastText and TextCNN. FastText works well on Chinese data~\cite{cmid} but not for English. RoBERTa gives the highest accuracy for the overall mean and each individual dataset.
\fi
%\subsection{Named Entity Recognition}

The evaluation outcomes are shown in Table \ref{results-table}. We find: 

\noindent \textbf{a) SFTs outperform LLMs:} We observe that all SFT approaches surpass ChatGPT by a big margin. Further, from Table~\ref{few-shot-table-binder}, it's evident that PubMedBERT can easily outperform ChatGPT on most benchmarks with just five supervised examples. 

\noindent \textbf{b) Transformer SFT Models:} i) PubMedBERT learns good embedding vectors due to the largest pretraining corpus. BINDER combined with PubMedBERT gives the best F1 score as it is able to leverage high-quality embeddings along with entity descriptions which pushes the similar entity tokens closer in the embedding space with a contrastive loss objective. (ii) LSTM/CNN-based neural embedding and traditional ML-based models - XGBoost and Logistic Regression perform poorly because they fail to capture contexts and do not leverage domain-specific pretraining.
%less context similarity.
%PubMedBERT also outperforms XLMR based approach \cite{xlmr-2022-naacl}. 
%From the results presented in \ref{result-table} we make following observations: \noindent \textbf{1.} Tranformer encoder based word-representation models outperform all other systems. Among those, PubMedBERT gives best performance because it is pre-trained on largest text corpus.\\
%(iii) Binder model is also outperforming XLMR based model~\cite{xlmr-2022-naacl}. 

\noindent \textbf{c) Feature-based SFT Models:} (i) ML-based model performs better than CNN/LSTM embedding systems on the DDI dataset, implying that it might be possible to beat the performances on other datasets if the right feature set is selected, which is usually an expensive process. (ii) The range of performance (best F1 - worst F1) for NCBI-Diease corpus is highest, showing that there is a big difference between the selected feature set and the features captured by the neural models. (iii) The addition of a CRF prediction layer on CNN/LSTM improves the performance significantly. 

\noindent \textbf{d) Dataset Quality:}  In most of the cases, low F1 is observed on the entities having fewer examples in the training set. For example, entities "CompositeMention" and "Disease Class" show poor performance due to less number of samples in training data. 
%More sample size will help to improve the performance.
%So, for BioMedical NER task, we can apply Binder(PubMedBERT). From Table \ref{experiments-joint-ner}, for Joint intent-entity extraction task, RoBERTa is performing best (except one case) so this model can be utilized for joint extraction.
%\textcolor{red}{more explain}\\
We note that the tag generation problem is difficult for instruction-tuned LLMs. %We observed very poor NER results on Llama-2 as it was unable to follow specified output structure and most of times, ended up hallucinating text. So, we omit LLaMa-2 results and leave adapter tuning comparison for future works. 
We also experiment with Llama-2 (7b) model%\footnote{https://ai.meta.com/llama/} 
\footnote{\href{https://ai.meta.com/llama/}{https://ai.meta.com/llama/}}
and observe that vanilla Llama-2-7b does not achieve good results as it was unable to follow the specified output structure and most of times, ended up hallucinating text. So, we omit vanilla Llama-2 results and will explore further in future. %leave adapter tuning comparison for future works.

%\noindent \textbf{Experimental Setup:}
\subsubsection{Experimental Setup}
We experiment on Tesla T4 16GB GPU, 6 Gbps clock cycle and GDDR5 memory. All experiments (entity extraction and intent detection) took $\sim$60 minutes for training. We fine-tune the models for a maximum of 20 epochs with a learning rate of 5e-5 with AdamW optimizer and 10\% warm-up steps. The batch size is 16. %Number of training steps per second is 2.67.
%We use NLTK, Spacy etc. Python libraries. 
Additional details are included in the GitHub Repository. %for PubMedBERT on NCBI Disease corpus was 2.67. 
%A working demo with all details are in Github page. %We use open source implementations of Fasttext and other models. Other experimental details for other models can found on the Github page.
%\mukur{Due to large memory requirements of BINDER model, on the same 16 GB T4 GPU batch size of 1 was used.}

\begin{figure}[t]
%\vspace{-2mm}
     \centering
     \begin{subfigure}[b]{0.49\columnwidth}
         \centering
         \includegraphics[width=\columnwidth]{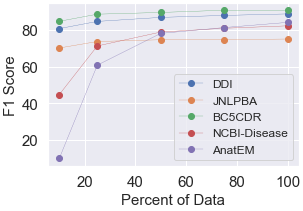}
         \caption{PubMedBERT}
         \label{fig2-1}
     \end{subfigure}
     \hfill
    % \begin{subfigure}[b]{0.49\columnwidth}
    %     \centering
    %     \includegraphics[width=\columnwidth]{data_ablation_bert.png}
    %     \caption{BERT}
    %     \label{fig2-2}
    % \end{subfigure}
    %  \hfill
     \begin{subfigure}[b]{0.49\columnwidth}
         \centering
         \includegraphics[width=\columnwidth]{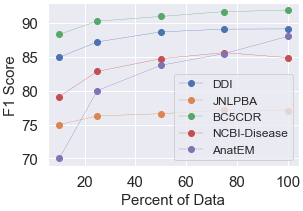}
         \caption{Binder-PubMedBERT}
         \label{fig2-3}
     \end{subfigure}
     \hfill
    % \begin{subfigure}[b]{0.49\columnwidth}
    %     \centering
    %     \includegraphics[width=\columnwidth]{data_ablation_lstm.png}
    %     \caption{LSTM+CRF}
    %     \label{fig2-4}
    % \end{subfigure}
    % \hfill
        %\caption{Increase in F1 score on test set by increasing the training data size.}
        %\vspace{-7mm}
        \caption{Ablation: Varying Training Size} %\textcolor{red}{remove b and d}}
        %\vspace{-4mm}
        \label{fig2}
\end{figure}

\begin{table}[b]
    %\vspace{-5mm}
    \centering
    \begin{adjustbox}{width=0.49\textwidth}
    \begin{tabular}{c|ccccc}
    \hline
    \textbf{\# Shots}        & \textbf{DDI}          & \textbf{JNLPBA}         & \textbf{BC5CDR}     & \textbf{NCBI}        & \textbf{AnatEM}  \\
    \hline
    5    &2.8 / 60.05                       & 2.0 / 39.09        & 1.1 / 64.53              & 3.2 / 27.56      & 2.6 / 1.5\\
    10     & 5.8 / 65.54             & 5.6 /50.56       & 2.1 /69.23                      & 7.05 /34.32    & 6.1 / 3.5\\
    30  & 45.49 /73.71           & 61.07 / 58.01     & 54.07 / 77.97           & 60.63 / 48.16   &21.74 / 10.9\\
    50     & 81.39 / 76.85             & 71.04 / 62.23       & 83.3 / 82.78      & 75.83 / 49.51     &40.69 / 30.96    \\
    100    & 83.56 / 80.94             & 74.24 / 68.02        & 88.58 / 85.76               & 82.38 / 72.78     & 73.34 / 52.73 \\
    \hline
    \end{tabular}
    \end{adjustbox}
    %\vspace{-2mm}
    \caption{BINDER-PubMedBERT / PubMedBERT F1-score in K-shot setting} %\textcolor{red}{one row, remove model column}} %Some results on AnatEm aren't shown due to very high variance in performance.}
    %\vspace{-2mm}
    \label{few-shot-table-binder} 
\end{table}

\section{Ablations}
We study the relationship of SFT models with domain-specific finetuning data:

\noindent \textbf{Varying Training Data Size:}  %{https://docs.google.com/spreadsheets/d/1foijvdvm4qJ4e2Qgh2CeyEA0wJIFMfNkYeuHEI-c3nY/edit?usp=sharing}{\color{red}{(Data Sheet link)}}} 
We vary the size of training data (10\%, 25\%, 50\%, 75\% and 100\%), while keeping the test set constant and show the performances of PubMedBERT, Binder (PubMedBERT) %, BERT, and LSTM+CRF 
models in Figure \ref{fig2}. We observe that, unlike raw PubMedBERT, Binder (PubMedBERT) attains a high performance with only 10\% of training data. Transformer-based models can learn with very little training data and performance does not decrease much even with 25\% training data. Due to domain pre-training, PubMedBERT learns with much fewer samples and saturates faster. This quick-learning behavior seems to be originating from transfer learning. %as we see the decrease in performance of PubMedBERT also. 
However, LSTM and CNN models suffer from poor performances in low-data settings due to no pretraining (details in GitHub). 

\noindent \textbf{Few Shot:} In Table \ref{few-shot-table-binder}, we show the performances (F1 Score) of the Binder (PubMedBERT) and PubMedBERT models on a few shot settings with different numbers of training samples. PubMedBERT embeddings perform better in very low-resource setups (5, 10 shots). However, when training examples increase further (30 shots onwards), BINDER (PubMedBERT) outperforms PubMedBERT because of the Bi-Encoder architecture trained on contrastive learning objectives. %We also note that raw PubMedBERT easily surpasses ChatGPT even in very less-resource setting (5 shots).
%It performs decently for a few samples in the case of drug and chemical compound (DDI and BC5CDR) NERs but more samples are required for disease/anatomical entity detection and perform very poorly for less data (first two cases). 

\begin{table}[!h]
%\vspace{-1mm}
\centering
\begin{adjustbox}{width=\linewidth}
\begin{tabular}{lccc}
\hline
\textbf{Error Type} & \textbf{Entity Text}  & \textbf{Label} & \textbf{Prediction}\\
\hline
Boundary & 3-[(...)ethynyl] pyridine & B-D,I-D & B-D,B-D\\
Entity type & heparinase III & B-D,I-D & B-G,I-G\\
Entity Miss & Hyaluronan lyase & B-D,I-D & O,O\\
\hline
\end{tabular}
\end{adjustbox}
%\vspace{-1.5mm}
\caption{Errors by BINDER-PubMedBERT on entity "drug\textunderscore n" of DDI dataset. %D:Drug and G:Group are abbreviated. 
Following abbreviations are used - B-D: B-DRUG\textunderscore N, I-D: I-DRUG\textunderscore N, B-G:B-Group, I-G: I-GROUP}
\label{DDI-error} 
%\vspace{-5mm}
\end{table}

\section{Error Analysis}
%Good F1 score of above 0.90 is observed on 3 out of 4 entities of DDI dataset by PubMedBERT. Very poor F1 score of 0.42 by PubMedBERT and 0.46 by BioBERT is seen on the entity "drug\textunderscore n" which represents new or unapproved drugs. 
We do a detailed analysis on errors as following: 

\noindent A) %Most of the errors in intent detection are with ``other'' category intent due overlapping pattern with different categories. 
Some errors are due to model failure like RoBERTa's failure to classify 52\% of the "other" intents from the KUAKE-QIC dataset. For example, a query such as ``I have a \textit{cyst} in the \textit{corner of my right eye} and it grows bigger and bigger.'' is classified wrongly as ``diagnosis'' intent but it is of ``other'' category. 

\noindent B) Three types of errors are observed for entity extraction (examples from the DDI dataset are shown in Table \ref{DDI-error}). 

\noindent
C) Some models fail to identify the entity ``drug\textunderscore n'' which represents new or unapproved drugs so a periodic model update is required. 

\noindent D) Relaxing entity-type error by considering exact F1-score instead of strict F1, we observe an uplift of 4.57\% in mean F1. % for PubMedBERT. %Some previous works like \citep{biobert} and others consider exact F1 which does not penalize the entity-type error, while some works use partial F1, which does not considers boundary prediction error. In our work, we use strict F1 which penalizes all three error type. 

\section{Conclusion}
The biomedical sector has matured significantly in the past few years. 
We show instead of relying on general-purpose LLMs, it is important to design an intent detection and entity extraction task for processing domain-specific texts. In this work, we show that fine-tuned RoBERTa and BINDER (PubMedBERT) can work efficiently to detect intents and extract named entities across various benchmark datasets in biomedical literature. In the future, we aim to extract intent and entity jointly as a relation tuple %the joint intent detection and entity extraction task for different domains. 
and inspect the performances of various cross-domain scenarios.

%\bibliography{aaai24}

%\end{document}

\section{Limitations}
Our dataset needs to be scaled up in terms of different languages, sizes, and intent labels which we aim to do in the near future. The approach needs to be updated as a single model for jointly extracting intents and entities for multilingual scenarios which we aim to do as a part of future work.

\section{Ethical Concerns}
We propose to release the algorithmic details and work on public datasets that neither reveal any personal sensitive information nor any toxic statement. So there are no ethical concerns in this work. %Besides, we have paid enough token money %(exact remuneration will be revealed once accepted to the conference)to the domain-expert annotators who have helped us in manually tagging the medical queries. %In accordance, 

\section{Acknowledgements}
 The project was supported in part by the grant given by I-Hub Foundation for Cobotics, IIT Delhi for the project, "Voice based Natural Interaction for Goal Oriented Tasks in Healthcare".

\section{Bibliographical References}
\bibliographystyle{lrec-coling2024-natbib}
\bibliography{main}

\end{document}